\documentclass[10pt,twocolumn,letterpaper]{article}

\usepackage{cvpr}
\usepackage{times}
\usepackage{epsfig}
\usepackage{graphicx}
\usepackage{tabularx}
\usepackage{amsmath}
\usepackage{amssymb}
\usepackage{bm}
\usepackage{multirow}
\usepackage{subcaption}
\newcolumntype{P}[1]{>{\centering\arraybackslash}p{#1}}

\usepackage[pagebackref=true,breaklinks=true,letterpaper=true,colorlinks,bookmarks=false]{hyperref}

\cvprfinalcopy 

\begin{document}

\title{Coarse-to-Fine Volumetric Prediction for Single-Image 3D Human Pose}

\author{Georgios Pavlakos$^1$, Xiaowei Zhou$^1$, Konstantinos G. Derpanis$^2$, Kostas Daniilidis$^1$ \\[0ex]
$^1$ University of Pennsylvania \hspace{2em} $^2$ Ryerson University\\ 
}

\maketitle

\begin{abstract}
This paper addresses the challenge of 3D human pose estimation from a single color image. Despite the general success of the end-to-end learning paradigm, top performing approaches employ a two-step solution consisting of a Convolutional Network (ConvNet) for 2D joint localization and a subsequent optimization step to recover 3D pose. In this paper, we identify the representation of 3D pose as a critical issue with current ConvNet approaches and make two important contributions towards validating the value of end-to-end learning for this task. First, we propose a fine discretization of the 3D space around the subject and train a ConvNet to predict per voxel likelihoods for each joint. This creates a natural representation for 3D pose and greatly improves performance over the direct regression of joint coordinates. Second, to further improve upon initial estimates, we employ a coarse-to-fine prediction scheme. This step addresses the large dimensionality increase and enables iterative refinement and repeated processing of the image features. The proposed approach outperforms all state-of-the-art methods on standard benchmarks achieving a relative error reduction greater than $30\%$ on average. Additionally, we investigate using our volumetric representation in a related architecture which is suboptimal compared to our end-to-end approach, but is of practical interest, since it enables training when no image with corresponding 3D groundtruth is available, and allows us to present compelling results for in-the-wild images.
\end{abstract}

\section{Introduction}
Estimating the full-body 3D pose of a human from a single monocular image is an open challenge, which has garnered significant attention since the early days of computer vision~\cite{lee1985determination}. Given its ill-posed nature, researchers have generally approached 3D human pose estimation in simplified settings, such as assuming background subtraction is feasible~\cite{agarwal2006recovering}, relying on groundtruth 2D joint locations to estimate 3D pose~\cite{ramakrishna2012,zhou20153d}, employing additional camera views~\cite{burenius20133d,kazemi2013multi}, and capitalizing on temporal consistency to improve upon single frame predictions~\cite{urtasun20063d,andriluka2010monocular}. This diversity of assumptions and additional information sources exemplifies the challenge presented by the task.

\begin{figure}[t!]
	  \centering
	  \includegraphics[width=1\linewidth,trim={3cm 8cm 0cm 5cm},clip]{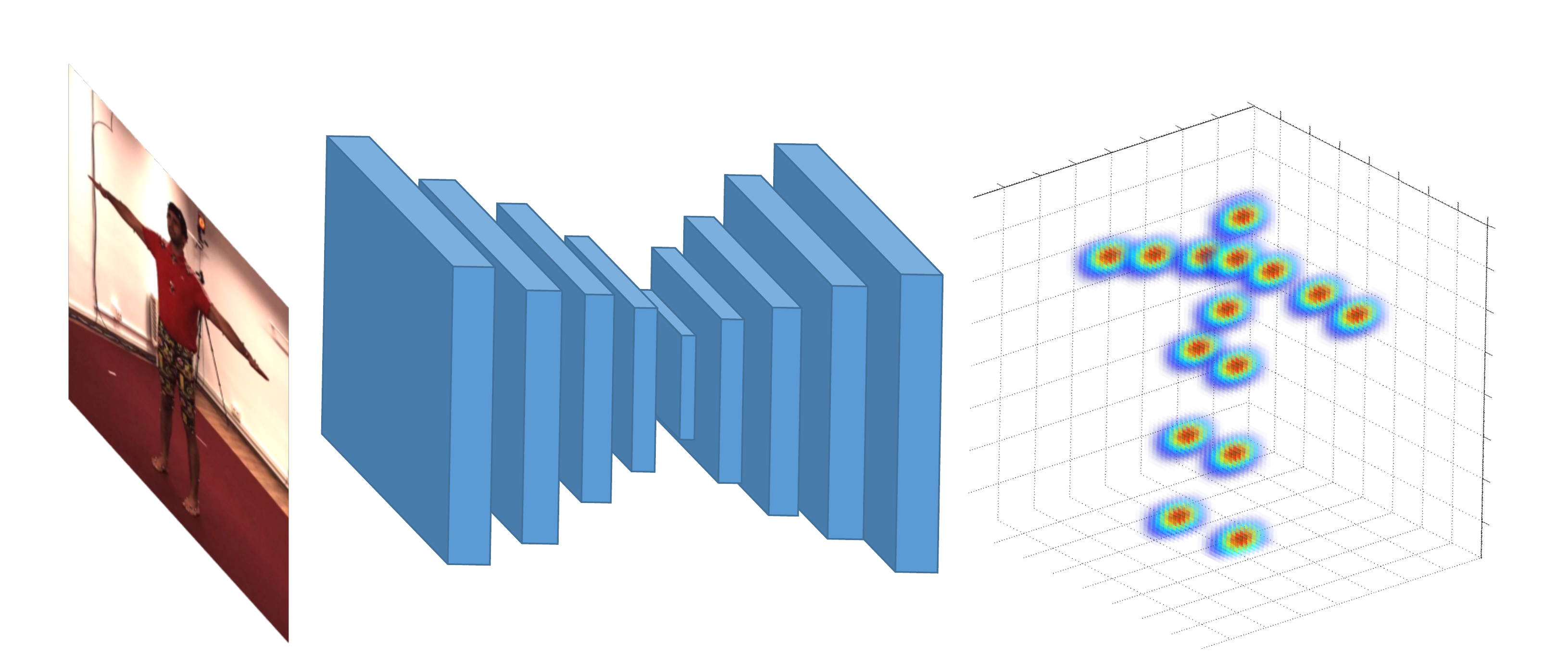}
	  \begin{tabular}{P{0.6cm} P{3.5cm} P{2cm}} 
	  \scriptsize{Image} & \scriptsize{ConvNet} & \scriptsize{Volumetric Output} \\ 
	  \end{tabular}
	  \vspace{-5pt}
    \caption{Illustration of our volumetric representation for 3D human pose. We discretize the space around the subject and use a ConvNet to predict per voxel likelihoods for each joint from a single color image.}
 	\label{fig:volumetric}
\end{figure}

With the introduction of more powerful discriminative approaches, such as Convolutional Networks (ConvNets), many of these restrictive assumptions have been relaxed. End-to-end learning approaches attempt to estimate 3D pose directly from a single image by addressing it as coordinate regression~\cite{li20143d, tekin2016structured}, nearest neighbor between images and poses~\cite{li2015maximum}, or classification over a set of pose classes~\cite{rogez2016mocap}. Yet to date, these approaches have been outperformed by more traditional two-step pipelines, e.g.,~\cite{zhou2016sparseness,bogo2016keep}. In these cases, ConvNets are used only for 2D joint localization and 3D poses are generated during a post-processing optimization step. Combining accurate 2D joint localization with strong and expressive 3D priors has been proven to be very effective. In this work, we show that ConvNets are able to provide much richer information than simply 2D joint locations.
	
To fully exploit the potential of ConvNets in the context of 3D human pose, we propose the following items, and justify them empirically.	First, we cast 3D pose estimation as a keypoint localization problem in a discretized 3D space. Instead of directly regressing the coordinates of the joints (e.g.,~\cite{li20143d, tekin2016structured}), we train a ConvNet to predict per voxel likelihoods for each joint in this volume. This volumetric representation, illustrated in Figure~\ref{fig:volumetric}, is much more sensible for the 3D nature of our problem and improves learning. Effectively, for every joint, the volumetric supervision provides the network with groundtruth for each voxel in the 3D space. This provides much richer information than a set of world coordinates. The empirical results also validate the superiority of our proposed form of supervision.
	
\begin{figure*}[htbp]
  \centering
  \includegraphics[width=1\linewidth,trim={2.7cm 9.5cm 3.5cm 7.4cm},clip]{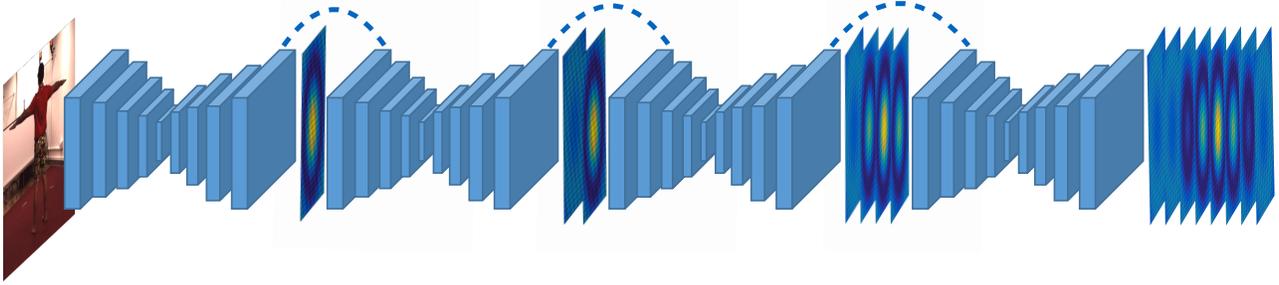}
  \vspace{-15pt}
  \caption{Illustration of our coarse-to-fine volumetric approach for 3D human pose estimation from a single image. The input is a single color image and the output is a dense 3D volume with separate per voxel likelihoods for each joint. The network consists of multiple fully convolutional components~\cite{newell2016stacked}, which are supervised in a coarse-to-fine fashion, to deal with the large dimensionality of our representation. 3D heatmaps are synthesized for supervision by increasing the resolution for the most challenging $z$-dimension (depth) after each component. The dashed lines indicate that the intermediate heatmaps are fused with image features to produce the input for the next fully convolutional component. For presentation simplicity, the illustrated heatmaps correspond to the location of only one joint.}
  \label{fig:coarse-to-fine}
\end{figure*}

Second, to deal with the increased dimensionality of the volumetric representation, we propose a coarse-to-fine prediction scheme. As demonstrated in the 2D pose case, intermediate supervision and iterative estimation are particularly effective strategies~\cite{wei2016cpm,carreira2016iterative,newell2016stacked}. For our volumetric representation though, naively stacking an increasing number of components and refining the estimates is not an effective solution, as shown empirically. Instead, we gradually increase the resolution of the supervision volume for the most challenging $z$-dimension (depth), during the processing. This coarse-to-fine supervision, illustrated schematically in Figure~\ref{fig:coarse-to-fine}, allows for more accurate estimates after each step. We empirically demonstrate the advantage of this practice over naively stacking more components together.
	
Our proposed approach achieves state-of-the-art results on standard benchmarks, outperforming both ConvNet-only  and hybrid approaches that post-process the 2D output of a ConvNet. Additionally, we investigate using our volumetric representation within a related architecture that decouples 2D joint localization and 3D joint reconstruction. In particular, we use two separate networks (the output of one serves as the input to the other) and two non-corresponding data sources, i.e., 2D labeled imagery to train the first component and an independent 3D data source (e.g., MoCap) to train the second one separately.  While this architecture has practical benefits (e.g., predicting 3D pose for in-the-wild images), we show empirically that it underperforms compared to our end-to-end approach when images with corresponding 3D groundtruth are available for training. This finding further underlines the benefit of predicting 3D pose directly from an image, whenever this is possible, instead of using 2D joint localization as an intermediate step.	

In summary, we make the four following contributions:
\begin{itemize}
\setlength\itemsep{0.0em}
  \item we are the first to cast 3D human pose estimation as a 3D keypoint localization problem in a voxel space using the end-to-end learning paradigm;
  \item we propose a coarse-to-fine prediction scheme to deal with the large dimensionality of our representation and enable iterative processing to realize further benefits;
  \item our proposed approach achieves state-of-the-art results on standard benchmarks, surpassing both ConvNet-only and hybrid approaches that employ ConvNets for 2D pose estimation, with a relative error reduction that exceeds $30\%$ on average;
  \item we show the practical use of our volumetric representation in cases when end-to-end training is not an option and present compelling results on in-the-wild images.
\end{itemize}

\section{Related work}
The literature on 3D human pose estimation is vast with approaches addressing the problem in a variety of settings. Here, we survey works that are most relevant to ours with a focus on ConvNet-based approaches; we refer the reader to a recent survey~\cite{sarafianos20163d} for a more complete literature review.
	 
The majority of recent ConvNet-only approaches cast 3D pose estimation as a coordinate regression task, with the target output being the spatial $x,y,z$ coordinates of the human joints with respect to a known root joint, such as the pelvis. Li and Chan~\cite{li20143d} pretrain their network with maps for 2D joint classification. Tekin \etal~\cite{tekin2016structured} include a pretrained autoencoder within the network to enforce structural constraints on the output. Ghezelghieh \etal~\cite{ghezelghieh2016learning} employ viewpoint prediction as a side task to provide the network with global joint configuration information. Zhou \etal~\cite{zhou2016deep} embed a kinematic model to guarantee the validity of the regressed pose. Park \etal~\cite{park20163d} concatenate the 2D joint predictions with image features to improve 3D joint localization. Tekin \etal~\cite{tekin2015direct} include temporal information in the joint predictions by extracting spatiotemporal features from a sequence of frames. In contrast to all these approaches, we adopt a volumetric representation of the human pose, and regress the per voxel likelihood for each joint separately. This proves to have significant advantages for the network performance and provides a richer output compared to the low-dimensional vector of joint coordinates.
	
An alternative approach to the classical regression paradigm is proposed by Li \etal~\cite{li2015maximum}. During training, they learn a common embedding between color images and 3D poses. At test time, the test image is coupled with each candidate pose and forwarded through the network; the input image is assigned the candidate pose with the maximum network score. This is a form of nearest neighbor classification which is highly inefficient due to the requirement of multiple forward network passes. On the other hand, Rogez and Schmid~\cite{rogez2016mocap} cast pose estimation as a classification problem. Given a predefined set of pose classes, each image is assigned to the class with the highest score. This guarantees a valid global pose prediction, but the approach is constrained by the poses in the original classes and thus returns only a rough pose estimate. In contrast to the inefficient nearest neighbor approach and the coarse classification approach, our volume regression allows for much more accurate 3D joint localization, while also being efficient.

Despite the interest in end-to-end learning, ConvNet-only approaches underperform those that employ a ConvNet for the 2D localization of joints, and produce 3D pose with a subsequent optimization step. Zhou \etal~\cite{zhou2016sparseness} utilize a standard 2D pose ConvNet to localize the joints and retrieve the 3D pose using an optimization scheme over a sequence of monocular images. Similarly, Du \etal~\cite{du2016marker} include height-maps of the human body to improve 2D joint localization. Bogo \etal~\cite{bogo2016keep} use the joints predicted by a 2D ConvNet and fit a statistical body shape model to recover the full shape of the human body. In contrast, our approach achieves state-of-the-art results with a single network. Furthermore, it provides a rich 3D output, amenable to post-processing, such as pictorial structures optimization to constrain limb lengths, or temporal filtering.
	
Another issue that has been addressed in the context of using ConvNets for 3D human pose is the scarcity of training data. Chen \etal~\cite{chen2016synthesizing} use a graphics renderer to create images with known groundtruth. Similarly, Ghezelghieh \etal~\cite{ghezelghieh2016learning} augment the training set with synthesized examples. A collage approach is proposed by Rogez and Schmid~\cite{rogez2016mocap}, where parts from in-the-wild images are combined to create additional images with known 3D poses. However, there is no guarantee that the statistics of the synthetic examples match those of real images. To investigate the data scarcity issue, we take inspiration from the 3D Interpreter Network~\cite{wu2016single}, which decouples the 3D pose estimation task into 2D localization and 3D reconstruction within a single ConvNet. In contrast, rather than using a predefined linear basis for 3D reconstruction, we predict 3D joint locations directly with our volumetric representation. This demonstrates the practical use of our volumetric representation even when end-to-end training is not an option. 

Finally, while we do not compare explicitly with multi-view pose estimation work (e.g.,~\cite{gall2010optimization,sigal2012,belagiannis2014,elhayek2015efficient}), it is interesting to note that the representation of 3D human pose in a discretized 3D space has also been previously adopted in multi-view settings~\cite{burenius20133d,kazemi2013multi,pavlakos2017harvesting}, where it was used to accommodate predictions from different viewpoints. For single view pose estimation, it has been considered in the context of random forests~\cite{kostrikov2014depth}. This approach suffered from large execution time (around three minutes), and required an additional refinement step using a pictorial structures model. In stark contrast, our network can provide complete volume predictions with a single forward pass in a few milliseconds, needs no additional refinement (although it is still a possibility) to provide state-of-the-art results, and is integrated within a coarse-to-fine prediction scheme to deal with excessive dimensionality.

\section{Technical approach}
The following subsections summarize our technical approach. Section \ref{sec:volumetric} describes the proposed volumetric representation for 3D human pose and discusses its merits. Next, Section \ref{sec:coarse2fine} describes our coarse-to-fine prediction approach that addresses the high dimensional nature of our output representation. Finally, Section \ref{sec:decoupled} describes the use of our volumetric representation within a related decoupled architecture and discusses its relative merits compared to our coarse-to-fine volumetric prediction approach.
\subsection{Volumetric representation for 3D human pose}\label{sec:volumetric}
The problem of 3D human pose estimation using ConvNets has been primarily approached as a coordinate regression problem. In this case, the target of the network is a $3N$-dimensional vector comprised of the concatenation of the $x,y,z$ coordinates of the $N$ joints of the human body. For training, an $\mathcal{L}_2$ regression loss is employed:
\begin{eqnarray}
  \mathcal{L} = \sum_{n=1}^N \| \bm{x}^n_{gt} - \bm{x}^n_{pr} \|_2^2,
\end{eqnarray}
where $\bm{x}^n_{gt}$ is the groundtruth and $\bm{x}^n_{pr}$ is the predicted location for joint $n$. The location of each joint is expressed globally, with respect to a root joint, or locally, with respect to its parent joint in the kinematic tree. The second formulation has some benefits, as discussed also by Li~\etal~\cite{li20143d} (e.g., easier to learn to predict small, local deviations), but still suffers from the fact that small errors can easily propagate hierarchically to children joints of the kinematic tree. In general, despite its simplicity, the coordinate regression approach makes the problem highly non-linear and presents problems for the learning procedure. These issues have previously been identified in the context of 2D human pose~\cite{tompson2014joint,pfister2015flowing}.

To improve learning, we propose a volumetric representation for 3D human pose. The volume around the subject is discretized uniformly in each dimension. For each joint we create a volume of size $w \times h \times d$. Let $p^n_{(i,j,k)}$ denote the predicted likelihood of joint $n$ being in voxel ${(i,j,k)}$. To train this network, the supervision is also provided in volumetric form. The target for each joint is a volume with a 3D Gaussian centered around the groundtruth position $\bm{x}_{gt}^n = {(x,y,z)}$ of the joint in the 3D grid:
\begin{eqnarray}
  G_{i,j,k}(\bm{x}_{gt}^n) = \frac{1}{2\pi \sigma^2} e^{-\frac{(x-i)^2 + (y-j)^2 + (z-k)^2}{2\sigma^2}},
\end{eqnarray}
where the value $\sigma = 2$ is used for our experiments. For training, we use the mean squared error loss:
\begin{eqnarray}
  \mathcal{L} = \sum_{n=1}^N \sum_{i,j,k} \|G_{(i,j,k)}(\bm{x}_{gt}^n) - p^n_{(i,j,k)}\|^2.
\end{eqnarray}
In theory, the output of the network is four dimensional, i.e., ($w \times h \times d \times N$), but in practice we organize it in channels, thus our output is three dimensional, i.e., $w \times h \times dN$. The voxel with the maximum response in each 3D grid is selected as the joint's 3D location.

A major advantage of the volumetric representation is that it casts the highly non-linear problem of direct 3D coordinate regression to a more manageable form of prediction in a discretized space. In this case, the predictions do not necessarily commit to a unique location for each joint, but instead an estimate of the confidence is provided for each voxel. This makes it easier for the network to learn the target mapping. A similar argument has been previously put forth in the 2D pose case, validating the benefit of predicting per pixel likelihoods instead of pixel coordinates~\cite{tompson2014joint,pfister2015flowing}. In terms of the network architecture, an important benefit of the volumetric representation is that it enables the use of a fully convolutional network for prediction. Here, we adopt the hourglass design \cite{newell2016stacked}. This leads to less network parameters than using fully connected layers for coordinate regression or pose classification. Finally, in terms of the predicted output, besides being more accurate, our network predictions in the form of dense 3D heatmaps are useful for subsequent post-processing applications. For example, structural constraints can be enforced with the use of a 3D Pictorial Structures model, e.g.,~\cite{burenius20133d,pavlakos2017harvesting}. Another option is to use the dense predictions in a filtering framework in cases where multiple input frames are available.
\subsection{Coarse-to-fine prediction}\label{sec:coarse2fine}
A design choice that has been particularly effective in the case of 2D human pose is the iterative processing of the network output~\cite{carreira2016iterative,wei2016cpm,newell2016stacked}. Instead of using a single component with a single output, the network is forced to produce predictions in multiple processing stages. These intermediate predictions are gradually refined to produce more accurate estimates. Additionally, the use of intermediate supervision on the ``earlier'' outputs allows for a richer gradient signal, which has been demonstrated empirically as an effective learning practice~\cite{lee2015deeply,szegedy2015going}.

Inspired by the success of iterative refinement in the context of 2D pose, we also consider a gradual refinement scheme. Empirically, we found that naively stacking multiple components yielded diminishing returns because of the large dimensionality of our representation. In fact, for the highest 3D resolution of $64 \times 64 \times 64$ with 16 joints, we would need to estimate the likelihood for more that 4 million voxels. To deal with this curse of dimensionality, we propose to use a coarse-to-fine prediction scheme. In particular, the first steps are supervised with lower resolution targets for the (most challenging and technically unobserved) $z$-dimension. Precisely, we use targets of size $64 \times 64 \times d$ per joint, where $d$ typically takes values from the set  $\{1,2,4,8,16,32,64\}$. An illustration of this supervision approach is given in Figure~\ref{fig:coarse-to-fine}.

This strategy makes training more effective, and allows us to benefit from stacking multiple components together without suffering from overfitting or dimensionality issues. Intuitively, easier versions of the task are presented to the network during the early stages of processing, and the complexity increases gradually. This postpones the harder decisions until the very end of the processing, when all the available information has been processed and consolidated.
\subsection{Decoupled architecture with volumetric target}\label{sec:decoupled}
To further show the versatility of the proposed volumetric representation, we also employ it in a scenario where end-to-end training is not an option. This is usually the case for in-the-wild images, where accurate, large-scale acquisition of 3D groundtruth is not feasible. Inspired by the 3D Interpreter Network \cite{wu2016single}, we decouple 3D pose estimation in two sequential steps consisting of predicting 2D keypoint heatmaps, followed by an inference step of the 3D joint positions with our volumetric representation. The first step can be trained with 2D labeled in-the-wild imagery, while the second step requires only 3D data (e.g., MoCap). Independently, each of these sources are abundantly available.

\begin{figure}[t]
	\begin{minipage}{1.00\textwidth}
	
	\begin{subfigure}{.5\textwidth}
	  \centering
	  \includegraphics[width=0.99\linewidth,trim={4cm 12cm 4cm 12cm},clip]{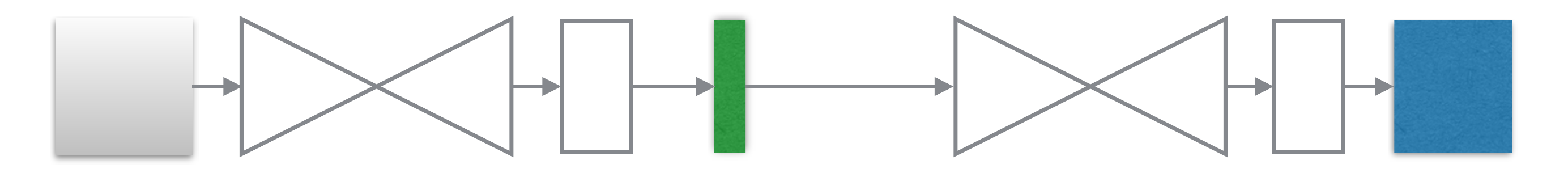}
	  \caption{Decoupled architecture}
	  \label{fig:rec2}
	\end{subfigure}
	
	\begin{subfigure}{0.5\textwidth}
	  \centering
	  \includegraphics[width=0.99\linewidth,trim={4cm 10cm 4cm 10cm},clip]{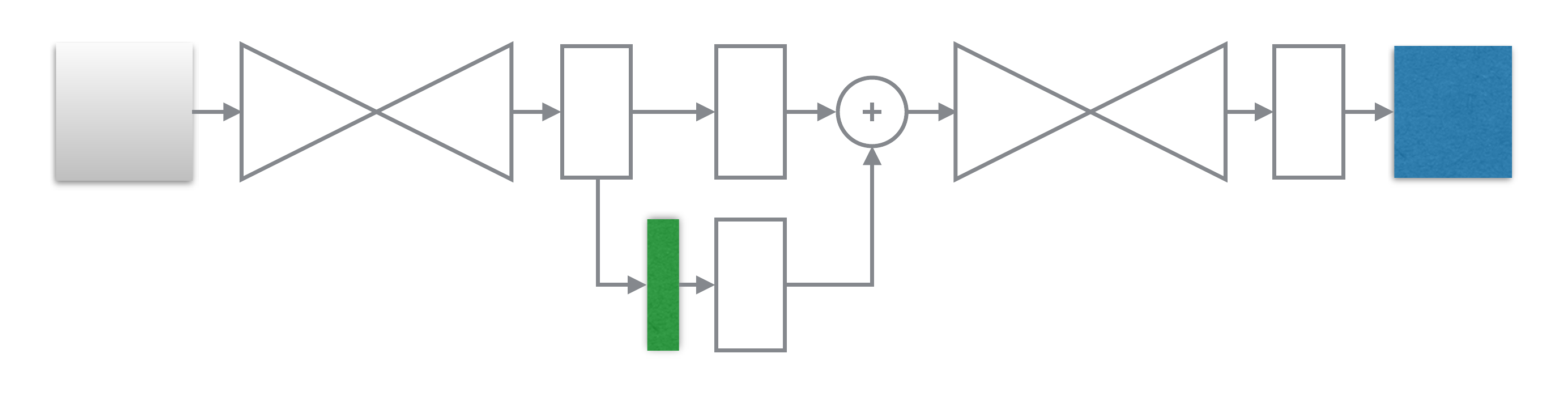}
	  \vspace{-0.5em}
	  \caption{Coarse-to-fine architecture}
	  \label{fig:rec1}
	\end{subfigure}
	
	\end{minipage}
    \caption{Schematic comparison of a decoupled architecture versus our coarse-to-fine architecture with intermediate supervision at the coarsest level (2D heatmaps). Blue blocks indicate 3D heatmaps, while green blocks indicate 2D heatmaps.
\textbf{Decoupled architecture}: The 2D heatmaps are provided directly as input to the second part of the network, which effectively operates as a 2D-to-3D reconstruction component. Note, no image features are processed in the second component, only information about 2D joint locations.
\textbf{Coarse-to-fine architecture}: We use 2D heatmaps as intermediate supervision, which are then combined with image features, effectively carrying information both from the image and the 2D locations of the joints.}
 \label{fig:decoupled}
\end{figure}

This training strategy is useful for practical scenarios, and we present compelling results for in-the-wild images (Sec.~\ref{sec:qualitative}). However, it remains suboptimal compared to our end-to-end approach when images with corresponding 3D groundtruth are available for training. Figure~\ref{fig:decoupled} provides an illustration of each architecture in a simplified setting with two hourglasses. It can be seen that the decoupled case is related to our course-to-fine architecture when the resolution of the intermediate supervision is set to $d = 1$ resulting in 2D heatmaps. A crucial difference between the two architectures is that our coarse-to-fine approach combines the produced 2D heatmaps with intermediate image features. This way, the rest of the network can process information both about the image and the 2D joint locations. On the other hand, a decoupled network processes the 2D heatmaps directly and attempts to reconstruct 3D locations without further aid by image-based evidence.  In cases where the heatmaps are grossly erroneous, the 3D predictions can be lead astray. In Sec.\ \ref{sec:component}, we show empirically that when images with corresponding 3D groundtruth are available, our coarse-to-fine architecture outperforms the decoupled one.

\section{Empirical evaluation}
\subsection{Datasets}
We present extensive quantitative evaluation of our coarse-to-fine volumetric approach on three standard benchmarks for 3D human pose: Human3.6M~\cite{ionescu2014human}, HumanEva-I~\cite{sigal2010humaneva} and KTH Football II~\cite{kazemi2013multi}. Additionally, qualitative results are presented on the MPII human pose dataset~\cite{andriluka20142d}, since no 3D groundtruth is available.

\vspace{3pt}
\noindent
\textbf{Human3.6M}: It contains video of 11 subjects performing a variety of actions, such as ``Walking'', ``Sitting'' and ``Phoning''. We follow the same evaluation protocol as prior work~\cite{li2015maximum,zhou2016sparseness}. In particular, Subjects S1, S5, S6, S7 and S8 were used for training, while subjects S9 and S11 were used for testing. The original videos were downsampled from 50fps to 10fps. We employed all camera views and trained a single model for \emph{all actions}, instead of training action-specific models \cite{li2015maximum,zhou2016sparseness}.

\vspace{3pt}
\noindent
\textbf{HumanEva-I}: It is a smaller dataset compared to Human3.6M, with fewer subjects and actions. Following the standard protocol~\cite{kostrikov2014depth,yasin2016dual}, we evaluated on ``Walking'' and ``Jogging'' from subjects S1, S2 and S3. The training sequences of these subjects and actions were used for training, and the corresponding validation sequences for testing. As done with the Human3.6M evaluation, we train a single model using the frames for all users and actions.

\vspace{3pt}
\noindent
\textbf{KTH Football II}: The images are taken from a professional football match, and 3D groundtruth is provided only for a very small number of them. The limited available groundtruth is not very accurate, since it was generated by combining manual 2D annotations from multiple views. In this case, image-to-3D training is not a practical option. Instead, we report results using our volumetric representation within the decoupled architecture described in Sec.~\ref{sec:decoupled}. More specifically, we train the first network component (image to 2D heatmaps) using images from this dataset which provide 2D groundtruth. For the second network component (2D heatmaps to 3D heatmaps), we use all the training MoCap data from Human3.6M dataset. As others~\cite{burenius20133d,tekin2015direct}, we report results using ``Sequence 1'' from ``Player 2'' and frames taken from ``Camera 1''.

\vspace{3pt}
\noindent
\textbf{MPII}: It is a large scale 2D pose dataset containing in-the-wild imagery. It provides 2D annotations but no 3D groundtruth. Like KTH, direct image-to-3D training is not a practical option with this dataset. Instead, we use the decoupled architecture with our volumetric representation. Since we cannot quantify the performance here, we only provide qualitative results.

\subsection{Evaluation metrics}
For Human3.6M, most approaches report the \textit{per joint 3D error}, which is the average Euclidean distance of the estimated joints to the groundtruth. This is done after aligning the root joints (here the pelvis) of the estimated and groundtruth 3D pose. An alternative metric, which is used by some methods to report results on Human3.6M and HumanEva-I is the \textit{reconstruction error}. It is defined as the per joint 3D error up to a similarity transformation. Effectively, the estimated 3D pose is aligned to the groundtruth by the Procrustes method. Finally, for KTH the percentage of the correctly estimated parts in 3D (3D PCP~\cite{burenius20133d}) is reported. Again, the root joints (here we use the center of the chest) are aligned to resolve the depth ambiguity.

\subsection{Implementation details}
In our volumetric space, the $x$-$y$ grid is a uniform discretization within the 2D bounding box in the image and the $z$ grid is a uniform discretization in $[-1,1]$ (meters) centered at the root joint. This means that we predict the image coordinates for each joint and its metric depth relative to the root. Given the depth of the root joint we can recover the absolute depth for each joint and its metric coordinates in the $x$-$y$ dimensions. For the component analysis (Section~\ref{sec:component}) we use the groundtruth depth of the root joint, while for the comparison to other methods (Section~\ref{sec:sota}) we estimate this depth based on the average skeleton size of each dataset. More details are provided in the supplementary material. 

In terms of the architecture, the fully convolutional components of our network, illustrated in Figure~\ref{fig:coarse-to-fine}, are based on the hourglass design~\cite{newell2016stacked}. We use the publicly available code to faithfully replicate the architecture. Similarly, we adopt the same training practices, employing rotation augmentation ($\pm30^o$), scaling augmentation ($0.75$-$1.25$), left-right flipping, and using rmsprop for optimization with the batch size equal to 4 and the learning rate set to $2.5\mathrm{e}{-4}$.

Regarding training on individual datasets, for Human3.6M, the network models are trained from scratch, typically for four epochs (approximately 310k iterations). For HumanEva-I, the model is trained from scratch for 120 epochs (approximately 235k iterations), because of the significantly smaller training set size. Finally, for the 2D joint localization network on KTH, we use the publicly available stacked hourglass model trained on MPII~\cite{newell2016stacked} and fine-tune it for 20 epochs (approximately 30k iterations).

\subsection{Component evaluation}\label{sec:component}
To evaluate the components of our approach, we use Human3.6M to report results, since it is the most complete available benchmark.

\vspace{3pt}
\noindent
\textbf{Volumetric representation}: Our first goal is to demonstrate that regression in a discretized space provides great benefit over coordinate regression. Both versions are implemented with the simplest setting of one hourglass. The only difference between the architectures is that the network for the volumetric predictions is fully convolutional, while for coordinate regression we use a fully connected layer at the end. The results are presented in Table~\ref{table:hm36m_volumetric}. The error of 112.41mm for coordinate regression is comparable to recent reported results with a coordinate regression target output~\cite{li20143d,tekin2016structured,park20163d,zhou2016deep}. In comparison, a significant decrease in the error is observed, down to 85.82mm at the highest depth resolution, by using the volumetric output target.

\begin{table}[t]
\begin{center}
\setlength{\tabcolsep}{2pt}
\begin{tabular}{lc}\cline{1-2}
 & Average \\\cline{1-2}
Coordinate Regression & 112.41 \\
Volume Regression ($d = 32$) & 92.23 \\
Volume Regression ($d = 64$) & \textbf{85.82} \\
\hline
\end{tabular}
\end{center}
\vspace{-15pt}
\caption{Coordinate versus volume regression on Human3.6M. The mean per joint error (mm) across all actions and subjects in the test set are shown.}
\label{table:hm36m_volumetric}
\end{table}

\vspace{3pt}
\noindent
\textbf{Coarse-to-fine prediction}: The next significant improvement to our network is provided by iterative processing of the image features. The naive way to achieve this is by stacking together multiple hourglasses. This is helpful, but offers only diminishing returns, as is demonstrated in Table~\ref{table:hm36m_coarse2fine} (Naive stacking). Instead, our coarse-to-fine supervision approach outperforms the naive one when we use two, three, or four hourglasses (Table~\ref{table:hm36m_coarse2fine}, Coarse-to-Fine). In fact, our coarse-to-fine version with two hourglasses (69.77mm error) outperforms the naive stacking network with four hourglasses (75.06mm error), despite using less than half of the parameters compared to the deeper network.\\
\begin{table}[t]
\begin{center}
\setlength{\tabcolsep}{2pt}
\begin{tabular}{|c|c|c|c|c||c|c|c|c|c|}\hline
\multicolumn{5}{|c||}{Naive Stacking} & \multicolumn{5}{c|}{Coarse-to-Fine}\\\hline
L1 	& L2 	& L3 	& L4 	& Avg.  & L1 	& L2 	& L3 	& L4 	& Avg. \\\cline{1-5}\hline \hline
\multirow{2}{*}{64} 	& \multirow{2}{*}{64}  	& \multirow{2}{*}{}  	& 	\multirow{2}{*}{} 	 	& \multirow{2}{*}{80.14}  & 1   	& 64 	& 			&			& 69.77 \\\cline{6-10}
& 			& 			& 			&  & 8   	& 64 	& 			&			& 77.52 \\\hline\hline
\multirow{2}{*}{64} 	& \multirow{2}{*}{64}  	& \multirow{2}{*}{64}  	& 	\multirow{2}{*}{} 	 	& \multirow{2}{*}{78.17}  & 1		&	2 		& 64		&			& 68.49 \\\cline{6-10}
& 			& 			& 			&  & 1		&	4 		& 64		&			& 72.02 \\\hline\hline
64 	& 64 	& 64 	& 64 	& 75.06 & 1		&	2 		& 4		&	64	& 64.76 \\\hline
\end{tabular}
\end{center}
\vspace{-15pt}
\caption{Comparison of the Naive Stacking (left) versus Coarse-to-Fine (right) approaches on Human3.6M. The column L$i$ denotes the $z$-dimension resolution for the supervision provided at the $i$-th hourglass component (empty if the network has less than $i$ components). We report mean per joint errors (mm) following the standard protocol.}
\label{table:hm36m_coarse2fine}
\end{table}

\vspace{3pt}
\noindent
\textbf{Decoupled network with volumetric representation}: 
We investigate the use of a decoupled network combined with our volumetric representation, as described in Sec.~\ref{sec:decoupled}. Our goal is to demonstrate the benefit of predicting the 3D pose directly from image features versus using 2D locations as an intermediate representation. We refer back to Fig.~\ref{fig:decoupled} for a schematic representation of the two relevant networks. We train both networks end-to-end to evaluate the architecture performance rather than the benefit of end-to-end training. (In fact we observed that training the two components of the decoupled network independently leads to inferior performance.) Comparative results are provided in Table~\ref{table:hm36m_decoupled}. We present the average across all actions, as well as the six actions with the largest difference between the two networks. Besides being more accurate for every action and on average, our coarse-to-fine approach shows significant improvement for more challenging actions, such as ``Sitting'' or ``Sitting Down''. In these cases, 2D joint localization often fails because of severe self-occlusions, providing the second subnetwork with inaccurate 2D heatmaps. Unless we process image features as well, 3D localization is bound to fail. This demonstrates the benefit of using information directly from the image for 3D localization versus decoupling the process.

\begin{table}[t]
\begin{center}
\setlength{\tabcolsep}{2pt}
\begin{tabular}{lcc}\hline
Action			&	Decoupled  &	Coarse-to-Fine \\ \hline
Phoning			&	75.00 		&	\textbf{66.50} 	\\ 
Sitting				&	95.25 		&	\textbf{76.99} 	\\ 
SittingDown		&	129.97 	&	\textbf{103.67}	\\ 
Smoking			&	75.58		&	\textbf{66.99}	\\ 
Walking			&	70.17		&	\textbf{59.12}	\\ 
WalkTogether	&	76.01		&	\textbf{62.28}	\\ 
Average			&	78.10		&	\textbf{69.77}	\\ \hline
\end{tabular}
\end{center}
\vspace{-15pt}
\caption{Comparison of our coarse-to-fine network using 2D heatmaps for intermediate supervision (Coarse-to-Fine) versus a decoupled network with a volumetric representation (Decoupled). The reported results are for the six classes of Human3.6M with the largest difference between the two approaches, as well as the average across all actions.}
\label{table:hm36m_decoupled}
\end{table}

\subsection{Comparison with state-of-the-art}\label{sec:sota}

\noindent
\textbf{Human3.6M}: We compare the performance of our approach with previously reported results on Human3.6M. Table~\ref{table:hm36m} presents the mean per joint 3D error results. Note that some previous works~\cite{tekin2015direct,zhou2016sparseness,du2016marker} leverage a sequence of frames for pose prediction rather than a single frame as considered by our approach. Nonetheless, our network achieves state-of-the-art results across the vast majority of actions and outperforms all other methods on average. Since some works use reconstruction error to report results, we also evaluate using this metric in Table~\ref{table:hm36m_align}. Again, our approach outperforms the other baselines by large margins.

\addtocounter{table}{1}
\begin{table}[]
\begin{center}
\setlength{\tabcolsep}{2pt}
\begin{tabular}{rcccc}\hline
& {\footnotesize Sanzari \etal~\cite{sanzari2016bayesian}} & {\footnotesize Rogez \etal~\cite{rogez2016mocap}} & {\footnotesize Bogo \etal~\cite{bogo2016keep}} & {\footnotesize Ours} \\\hline
{\small Average} & 93.2 & 87.3 & 82.3 & {\bf 51.9} \\
\hline
\end{tabular}
\end{center}
\vspace{-15pt}
\caption{Quantitative comparison on Human3.6M among approaches that report reconstruction error (mm). Baseline numbers are taken from the respective papers.}
\label{table:hm36m_align}
\end{table}

\vspace{3pt}
\noindent
\textbf{HumanEva-I}: Our empirical results for HumanEva-I are presented in Table~\ref{table:humaneva}, along with reported results from state-of-the-art approaches. Similar to Human3.6M, our network outperforms all other published approaches.

\begin{table}[t]
\small
\begin{center}\setlength{\tabcolsep}{2pt}
\begin{tabular}{r|ccc|ccc|c}\cline{1-8}
& \multicolumn{3}{c|}{Walking} & \multicolumn{3}{c|}{Jogging} & \multirow{2}{*}{Avg} \\
\cline{2-7}
& S1 & S2 & S3 & S1 & S2 & S3 & \\
\hline
Radwan \etal \cite{radwan2013monocular} & 75.1 & 99.8 & 93.8 & 79.2 & 89.8 & 99.4 & 89.5 \\
Wang \etal \cite{wang2014robust} & 71.9 & 75.7 & 85.3 & 62.6 & 77.7 & 54.4 & 71.3 \\
Simo-Serra \etal \cite{simo2013joint} & 65.1 & 48.6 & 73.5 & 74.2 & 46.6 & 32.2 & 56.7\\
Bo \etal \cite{bo2010twin} & 46.4 & 30.3 & 64.9 & 64.5 & 48.0 & 38.2 & 48.7 \\
Kostrikov \etal \cite{kostrikov2014depth} & 44.0 & 30.9 & 41.7 & 57.2 & 35.0 & 33.3 & 40.3 \\
Yasin \etal \cite{yasin2016dual} & 35.8 & 32.4 & 41.6 & 46.6 & 41.4 & 35.4 & 38.9 \\
Ours & \textbf{22.3} & \textbf{19.5} & \textbf{29.7} & \textbf{28.9} & \textbf{21.9} & \textbf{23.8} & \textbf{24.3}\\
\hline
\end{tabular}
\end{center}
\vspace{-15pt}
\caption{Quantitative results on HumanEva-I. The numbers are the mean reconstruction errors (mm). Baseline numbers are taken from the respective papers.}
\label{table:humaneva}
\end{table}

\vspace{3pt}
\noindent
\textbf{KTH Football II}: We report results for our approach on this dataset in Table~\ref{table:kth}, comparing with relevant methods. Note, Tekin \etal~\cite{tekin2015direct} use video for prediction instead of a single frame, while Burenius \etal~\cite{burenius20133d} is a multi-view method. Despite this, we outperform the single view approaches and we are competitive with the multi-view results.

\begin{table}
\centering
\begin{tabular}{lcc|ccc}\cline{1-6}
 & \cite{burenius20133d} & \cite{burenius20133d} & \cite{burenius20133d} & \cite{tekin2015direct} & Ours \\
\hline
Cameras & 3 & 2 & 1 & 1 (video) & 1 \\
\hline
Upper Arms & 60 & 53 & 14 & 74 & \textbf{96} \\
Lower Arms & 35 & 28 & 06 & 49 & \textbf{83} \\
Upper Legs & 100 & 88 & 63 & \textbf{98} & \textbf{98} \\
Lower Legs & 90 & 82 & 41 & 77 & \textbf{88} \\\cline{1-6}
\end{tabular}
\vspace{-5pt}
\caption{Quantitative results on KTH Football II. The numbers are the mean PCP scores (the higher the better). Baseline numbers are taken from the respective papers. We indicate how many cameras each approach uses, and highlight the best performance for single view approaches.
}
\label{table:kth}
\end{table}

\subsection{Qualitative results}\label{sec:qualitative}
\begin{figure*}[p]
\includegraphics[width=0.33\textwidth]{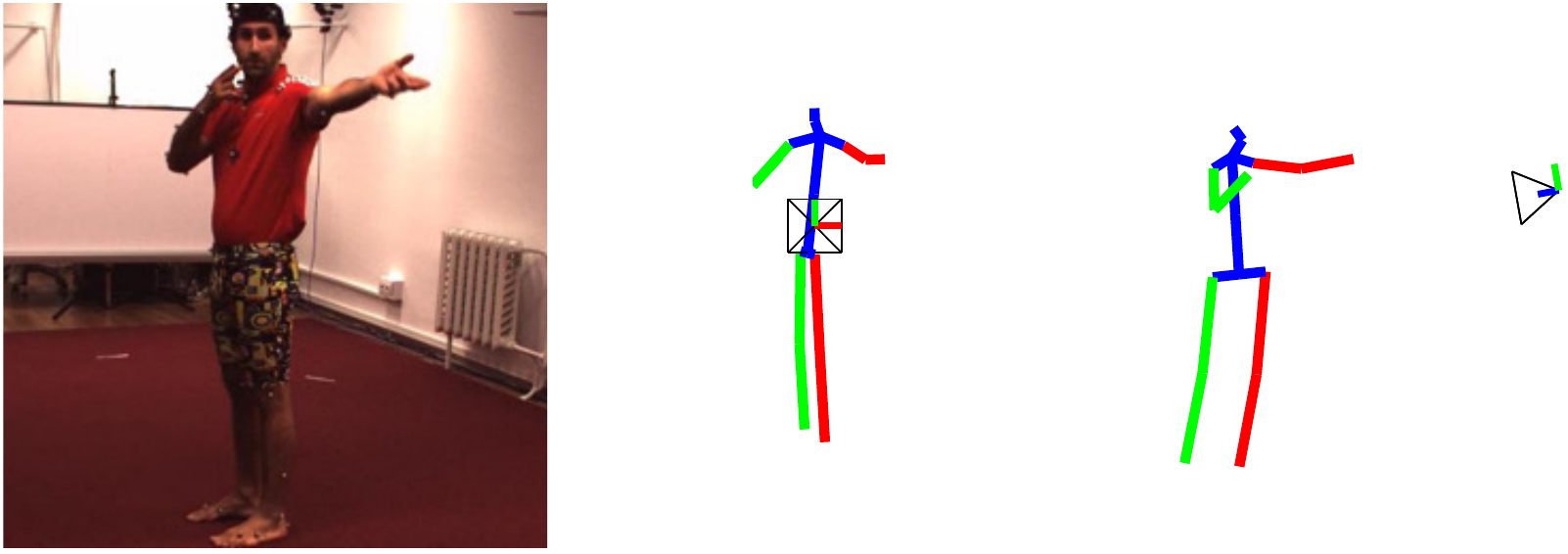}
\includegraphics[width=0.33\textwidth]{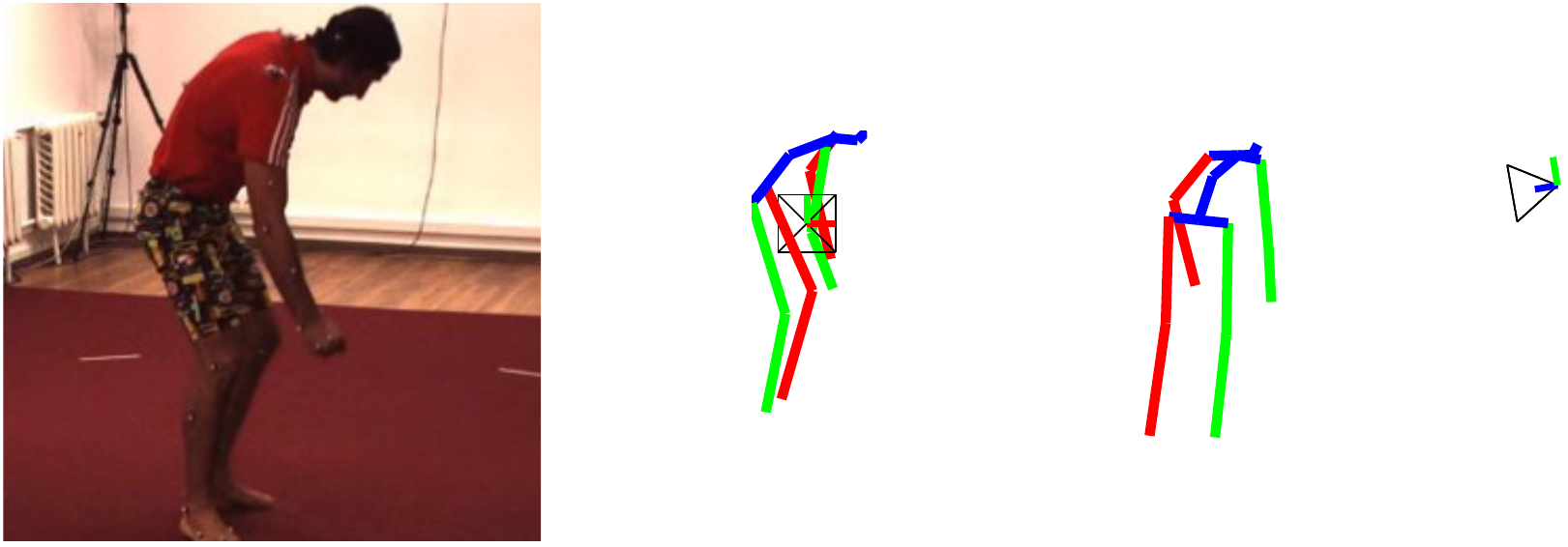}
\includegraphics[width=0.33\textwidth]{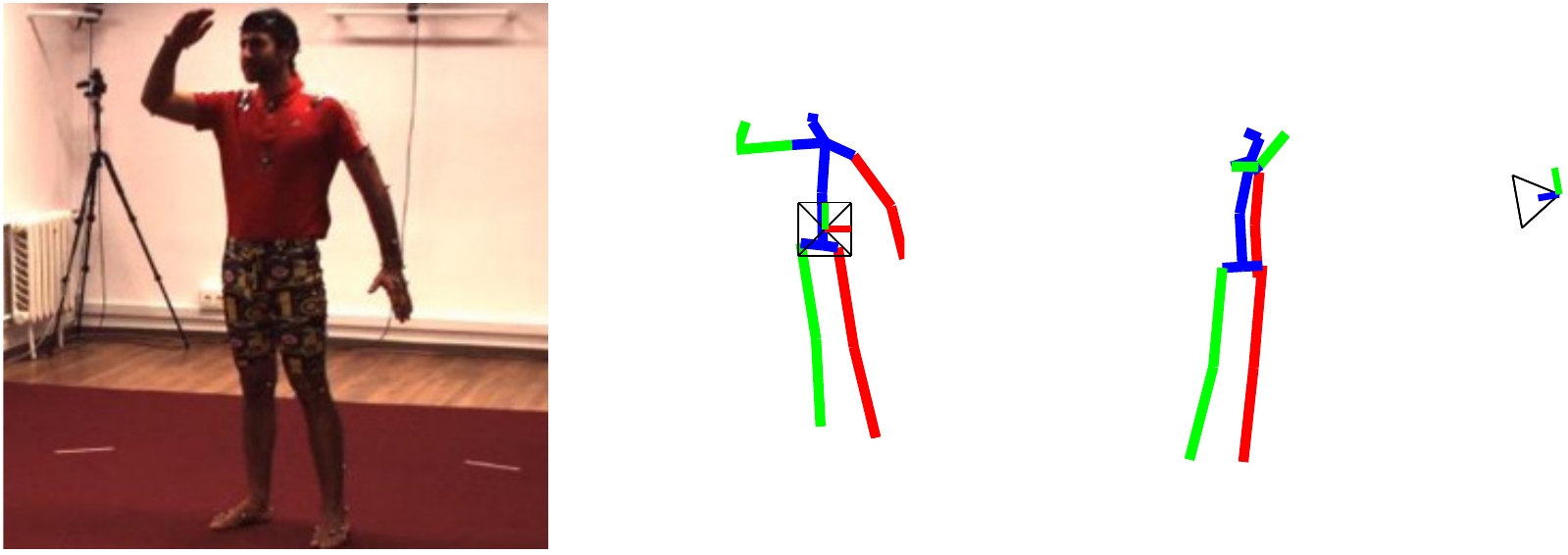}\vspace{0.01em}
\includegraphics[width=0.33\textwidth]{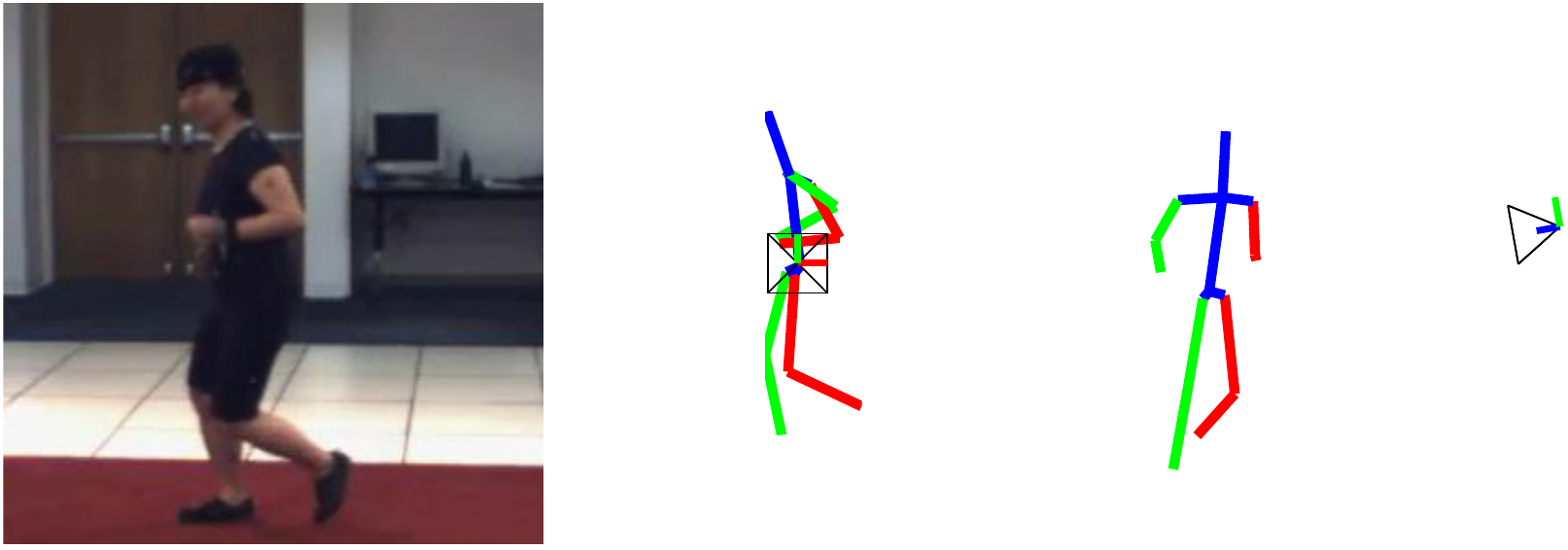}
\includegraphics[width=0.33\textwidth]{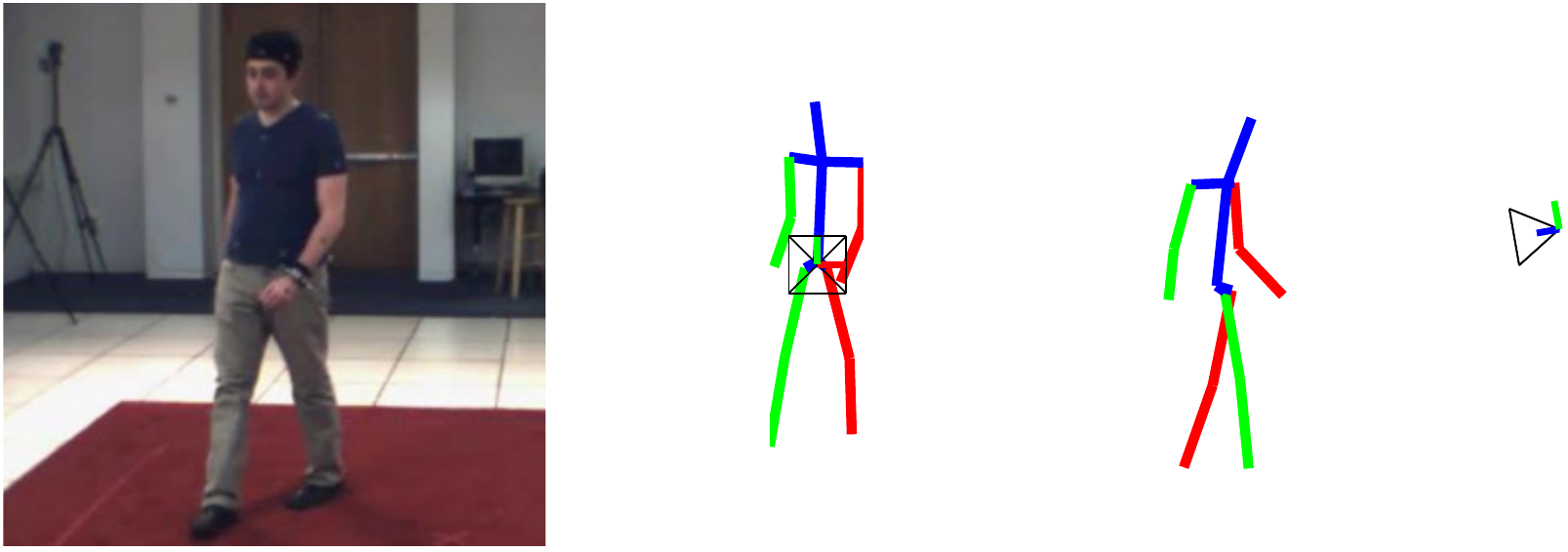}
\includegraphics[width=0.33\textwidth]{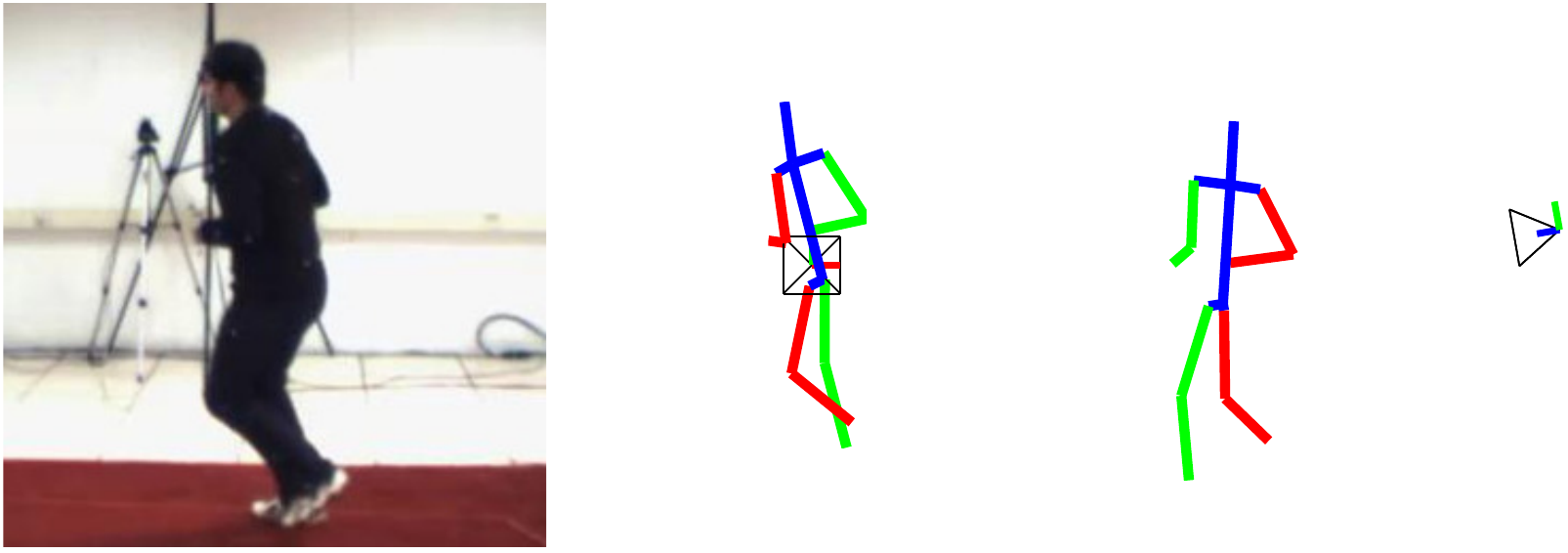} \vspace{0.01em}
\includegraphics[width=0.33\textwidth]{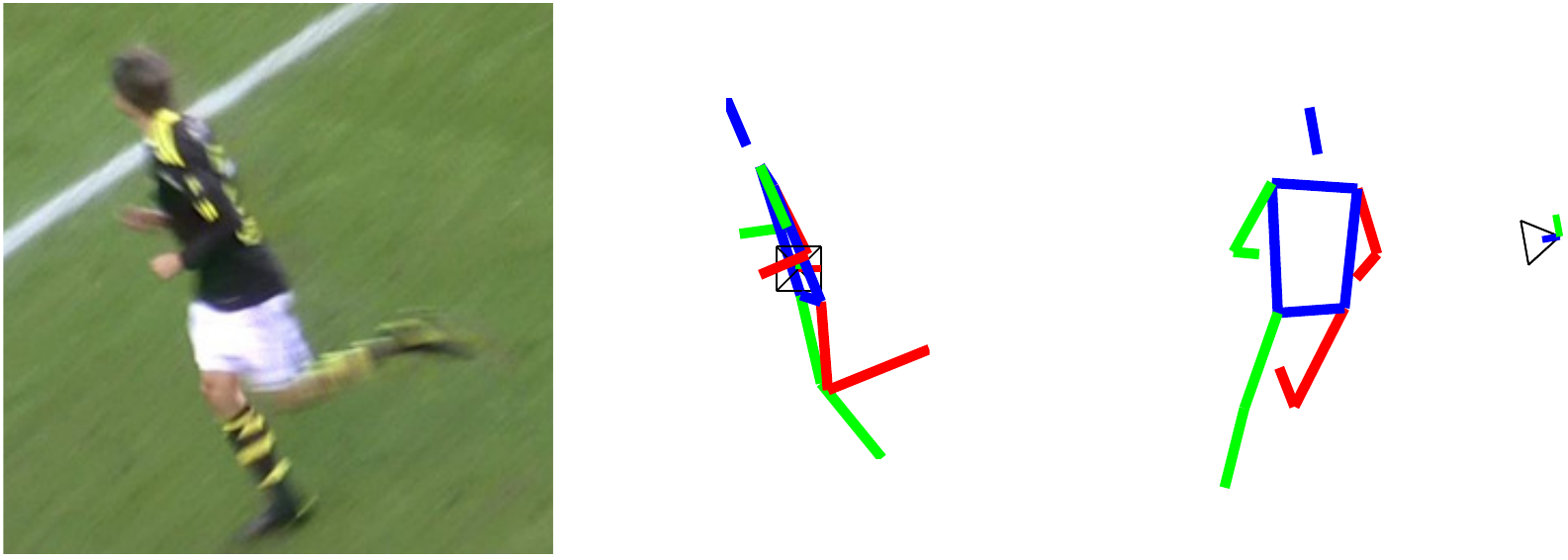}
\includegraphics[width=0.33\textwidth]{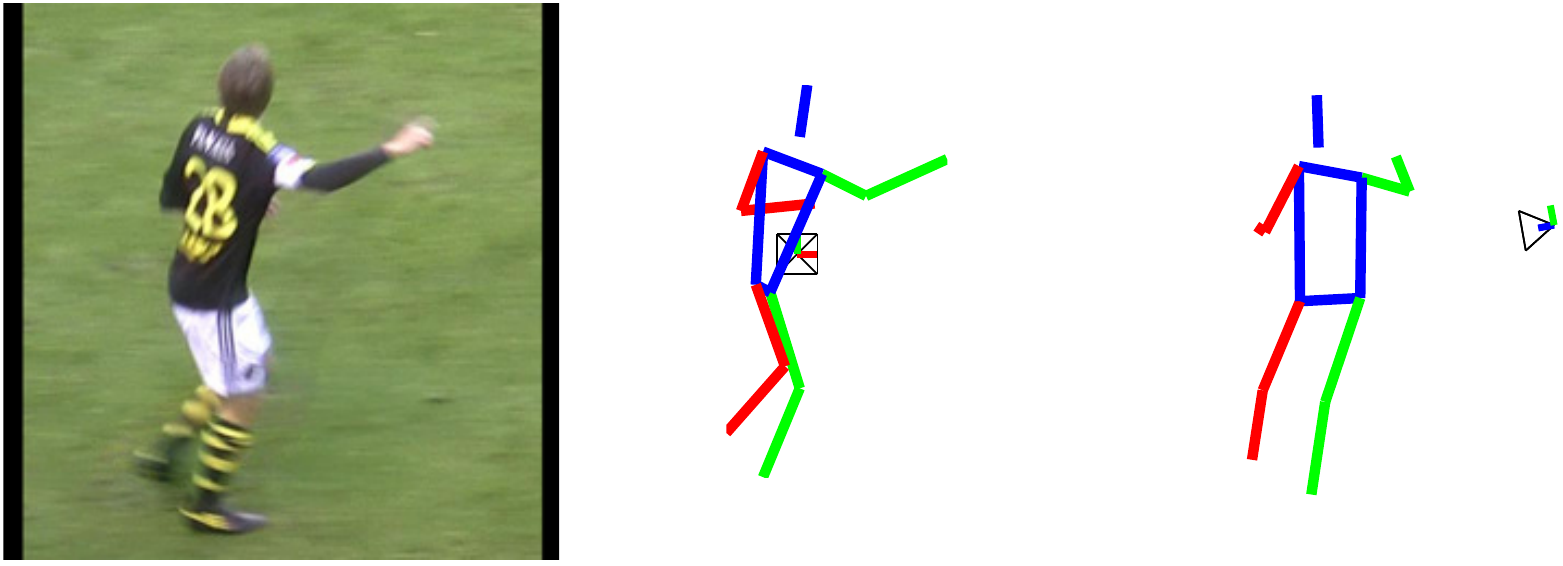}
\includegraphics[width=0.33\textwidth]{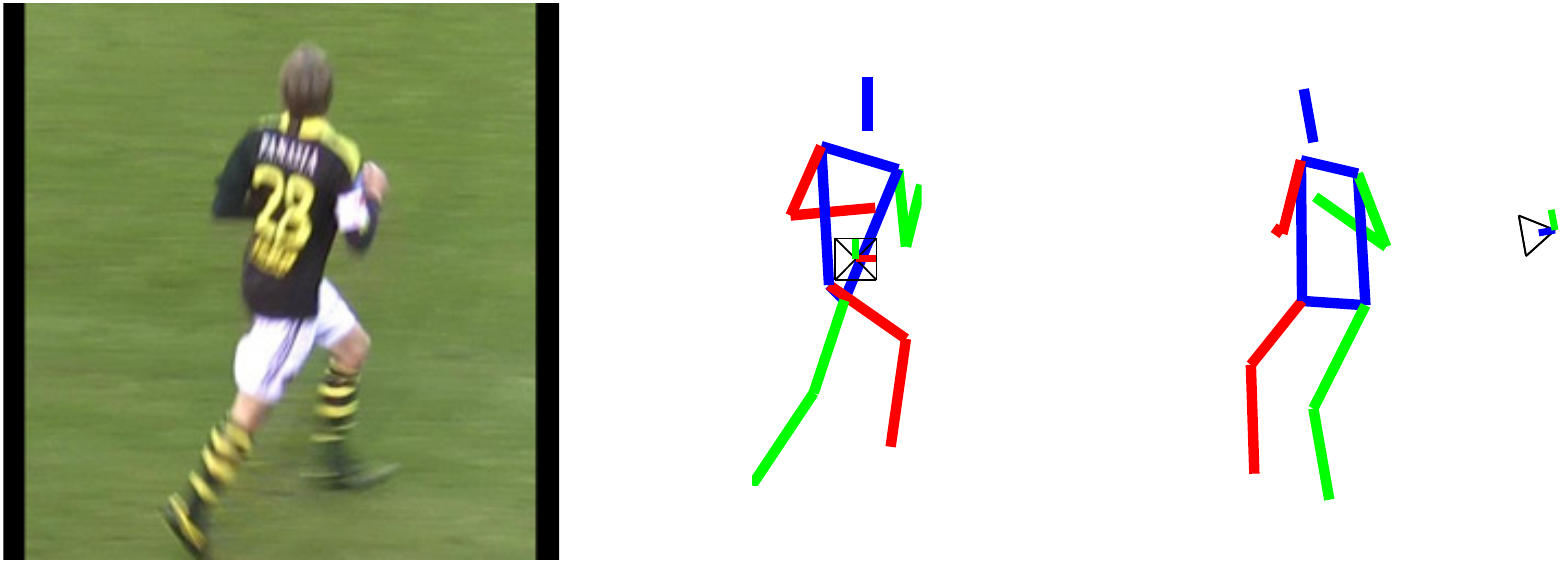} \vspace{0.01em}
\includegraphics[width=0.33\textwidth]{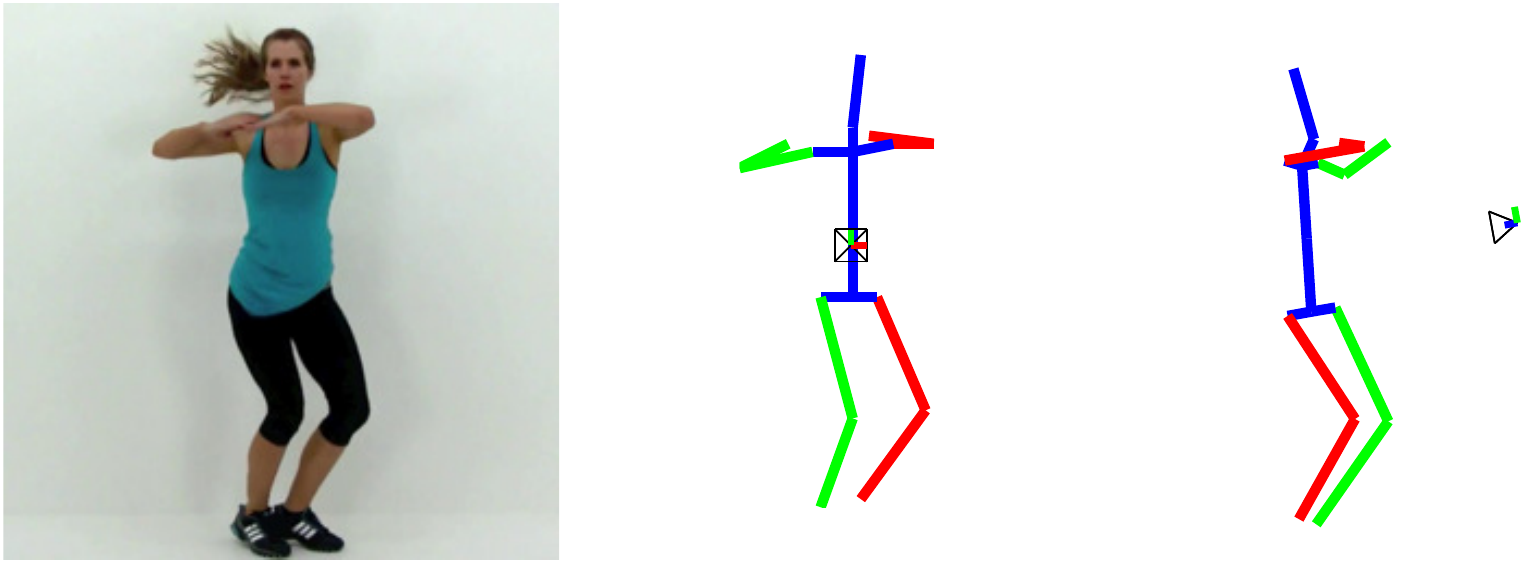}
\includegraphics[width=0.33\textwidth]{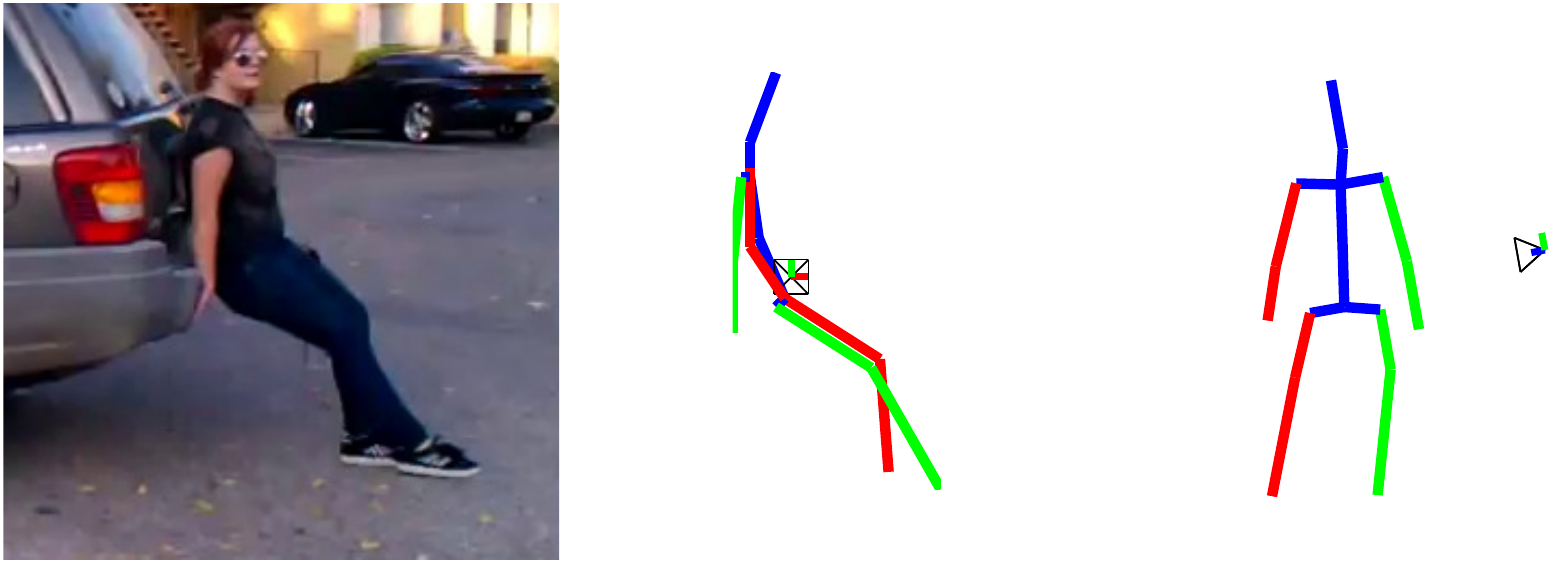}
\includegraphics[width=0.33\textwidth]{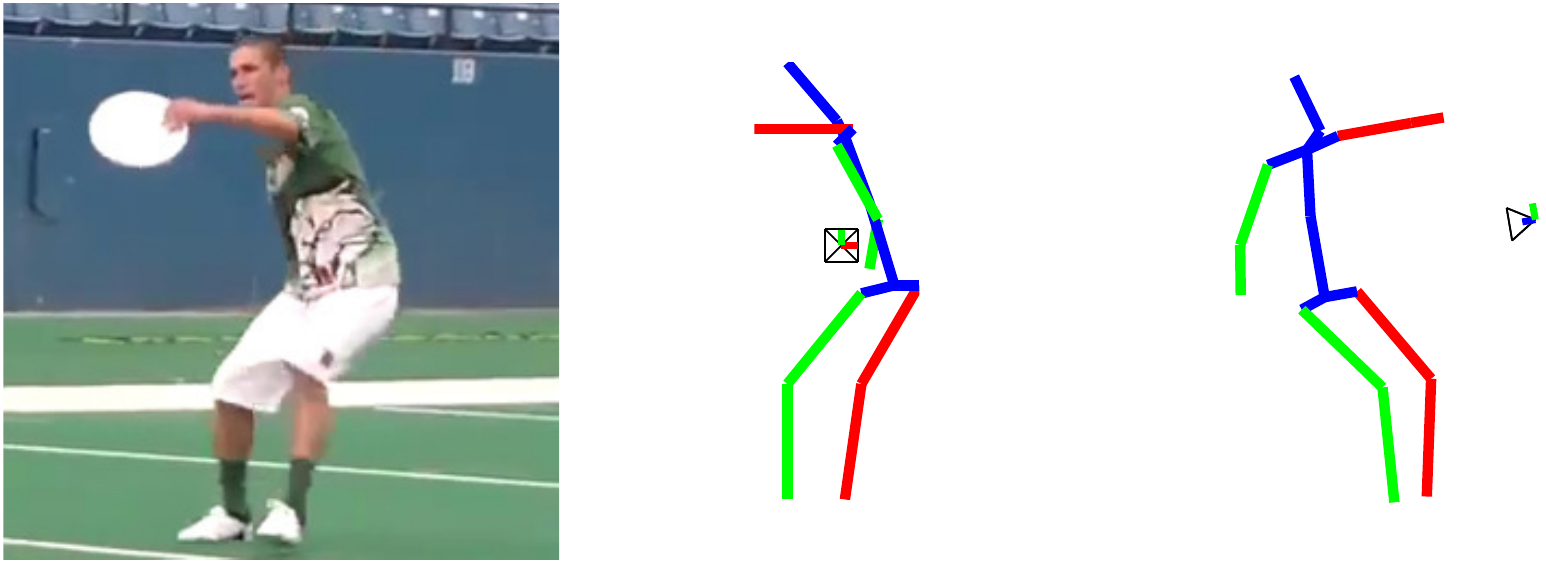} \vspace{0.01em}
\includegraphics[width=0.33\textwidth]{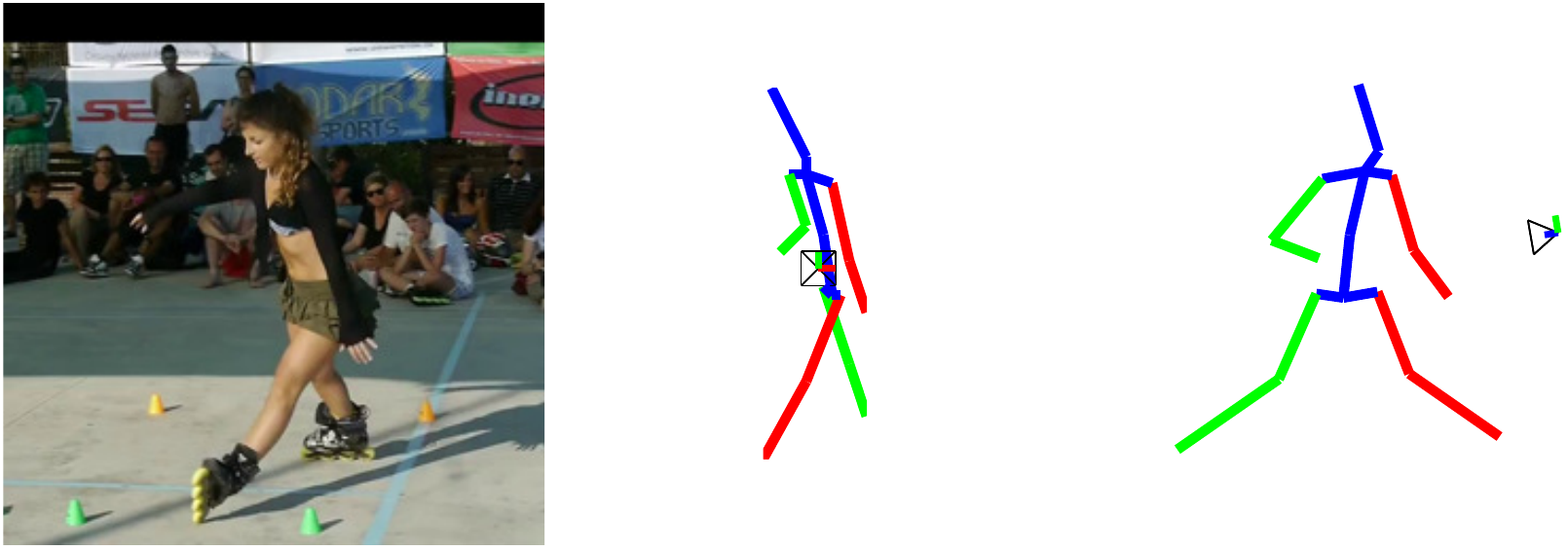}
\includegraphics[width=0.33\textwidth]{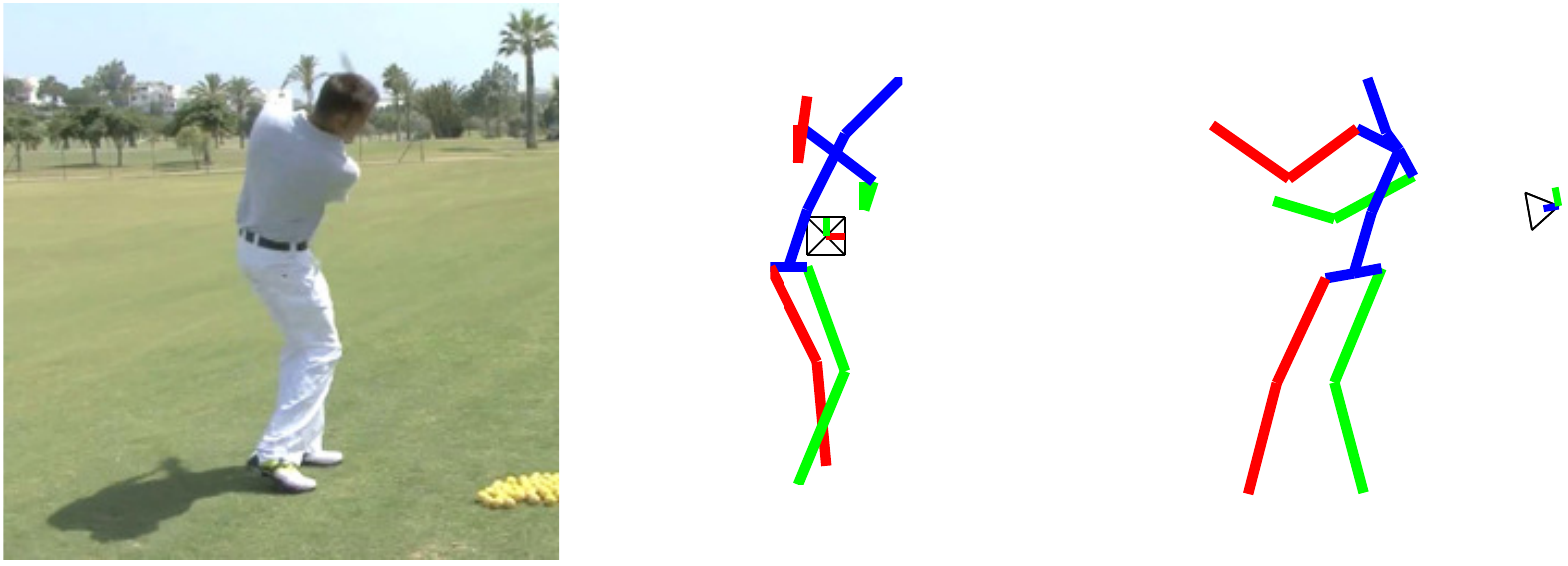}
\includegraphics[width=0.33\textwidth]{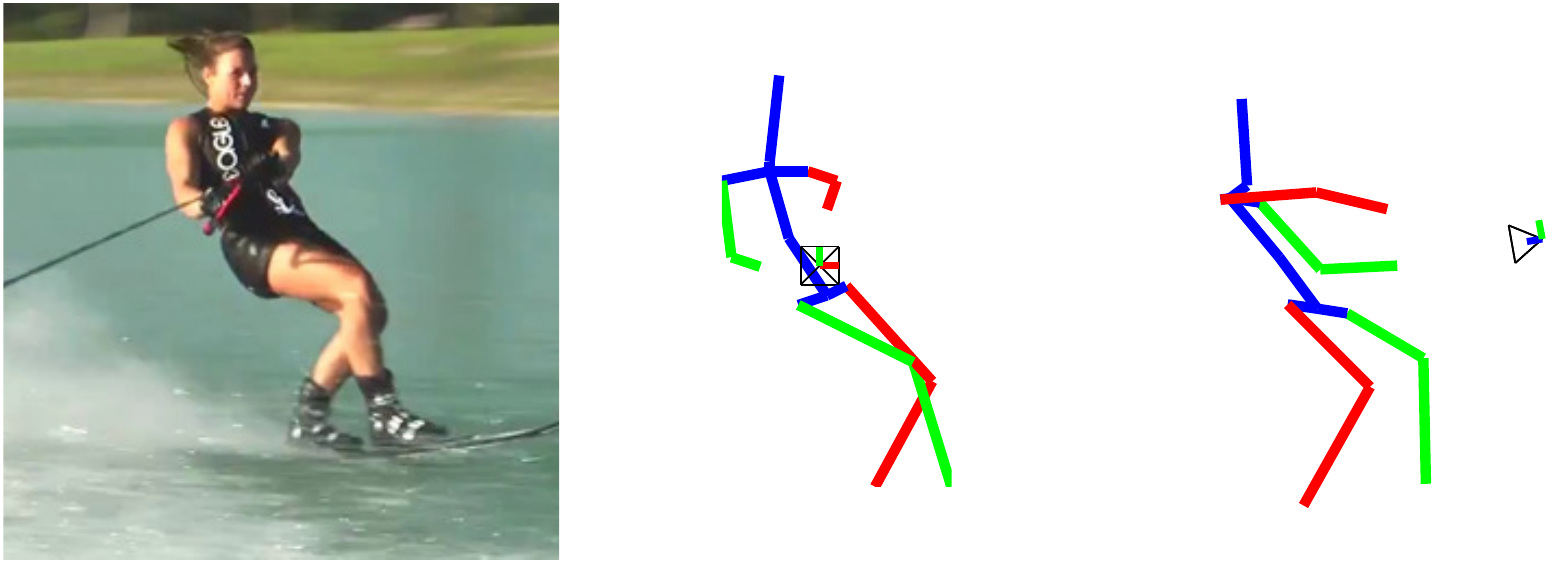}
\vspace{-10pt}
\caption{Sample qualitative results for all the datasets used in the empirical evaluation. First row: Human3.6M. Second row: HumanEva. Third row: KTH Football II. Fourth and fifth row: MPII. For each image, the original viewpoint and a novel viewpoint are shown. Red and green indicate left and right, respectively.}
\label{fig:qualitative}
\end{figure*}

Figure~\ref{fig:qualitative} presents qualitative results for images taken from the aforementioned datasets. Additionally, we demonstrate 3D reconstructions on MPII which offers greater visual variety due to its in-the-wild nature. Despite the challenging poses present in the MPII examples, our volumetric representation produces compelling 3D predictions.

\addtocounter{table}{-4}
\begin{table*}[hp]
\small
\begin{center}\setlength{\tabcolsep}{2pt}
\begin{tabular}{rcccccccc}\cline{1-9}
& Directions & Discussion & Eating & Greeting & Phoning & Photo & Posing & Purchases \\\hline
 LinKDE \cite{ionescu2014human} & 132.71 & 183.55 & 132.37 & 164.39 & 162.12 & 205.94 & 150.61 & 171.31 \\
Li et al. \cite{li2015maximum} & - & 134.13 & 97.37 & 122.33 & - & 166.15 & - & - \\ 
 Tekin \etal \cite{tekin2015direct} &        102.41 &        147.72 &       88.83 &        125.28 &        118.02 &        182.73 & 112.38 &        129.17 \\
 Zhou \etal \cite{zhou2016sparseness} &        87.36 &        109.31 &        87.05 &        103.16 &        116.18 &        143.32 &        106.88 &        99.78 \\
  Tekin \etal \cite{tekin2016structured}  &        - &        129.06 &        91.43 &        121.68 &        - &        162.17 &        - &        - \\
  Ghezelghieh \etal \cite{ghezelghieh2016learning}  &        80.30 &        80.39 &        78.13 &        89.72 &        - &        - &        - &        - \\
  Du \etal \cite{du2016marker}  &        85.07 &       112.68 &        104.90 &        122.05 &        139.08 &        135.91 &        105.93 &        166.16 \\
  Park \etal \cite{park20163d}  &         100.34 &        116.19 &        89.96 &        116.49 &        115.34 &        149.55 & 117.57 &        106.94 \\  
  Zhou \etal \cite{zhou2016deep}  &        91.83 &        102.41 &        96.65 &        98.75 &        113.35 &        125.22 &       90.04 &        93.84 \\
 Ours &        {\bf 67.38} &        {\bf 71.95} &        {\bf 66.70} &        {\bf 69.07} &        {\bf 71.95} &        {\bf 76.97} &       {\bf 65.03} &        {\bf 68.30} \\
\hline
\hline
&  Sitting & SittingDown &  Smoking &  Waiting & WalkDog &  Walking & WalkTogether &  Average \\\hline
 LinKDE \cite{ionescu2014human} & 151.57 & 243.03 & 162.14 & 170.69 & 177.13 & 96.60 & 127.88 & 162.14  \\
Li et al. \cite{li2015maximum} & - & - & - & - & 134.13 & 68.51 & - & - \\
  Tekin \etal \cite{tekin2015direct}  &        138.89 &        224.9 &        118.42 &       138.75 &        126.29 &       {\bf 55.07} &        65.76 &        124.97 \\
  Zhou \etal \cite{zhou2016sparseness}  &        124.52 &        199.23 &        107.42 &        118.09 &        114.23 &        79.39 &        97.70 &        113.01 \\
  Tekin \etal \cite{tekin2016structured}  &        - &       - &        - &        - &        130.53 &       65.75 &        - &        - \\
  Ghezelghieh \etal \cite{ghezelghieh2016learning} &        - &       - &        - &        - &        - &        95.07 &        82.22 &        - \\
  Du \etal \cite{du2016marker}  &        117.49 &       226.94&        120.02 &        117.65 &        137.36 &        99.26 &        106.54 &        126.47 \\
  Park \etal \cite{park20163d}  &         137.21 &        190.82 &        105.78 &        125.12 &       131.90 &        62.64 &        96.18 &        117.34 \\  
  Zhou \etal \cite{zhou2016deep}  &        132.16 &        158.97 &        106.91 &        94.41 &        126.04 &        79.02 &        98.96 &        107.26 \\
  Ours &        {\bf 83.66} &        {\bf 96.51} &        {\bf 71.74} &        {\bf 65.83} &        {\bf 74.89} &        59.11 &        {\bf 63.24} &        {\bf 71.90} \\
\hline
\end{tabular}
\end{center}
\vspace{-10pt}
\caption{Quantitative comparison on Human3.6M. The numbers are the average 3D joint error (mm). Baseline numbers are taken from the respective papers. Note, several approaches use video for prediction rather than a single frame \cite{tekin2015direct,zhou2016sparseness,du2016marker}.
}
\label{table:hm36m}
\end{table*}

\section{Summary}
Our paper addressed the challenging problem of 3D human pose estimation from a single color image. Departing from recent ConvNet approaches, we cast the problem as 3D keypoint localization in a discretized space around the subject. We integrated this volumetric representation with a coarse-to-fine supervision scheme to deal with the high dimensionality and enable iterative processing. We demonstrated that our contributions were crucial to achieve state-of-the-art results on the standard benchmarks with a relative error reduction greater than $30\%$ on average. Furthermore, we used our volumetric representation within a decoupled architecture, making it of practical use for in-the-wild images even when end-to-end training is not feasible.

\vspace{1em}
\footnotesize
\noindent
{\bf Project Page:} \url{https://www.seas.upenn.edu/~pavlakos/projects/volumetric}

\vspace{0.5em}
\footnotesize
\noindent
{\bf Acknowledgements:} We gratefully appreciate support through the following grants: NSF-DGE-0966142 (IGERT), NSF-IIP-1439681 (I/UCRC), NSF-IIS-1426840, ARL MAST-CTA W911NF-08-2-0004, ARL RCTA W911NF-10-2-0016, ONR N00014-17-1-2093, an ONR STTR (Robotics Research), NSERC Discovery, and the DARPA FLA program.

{\small
\bibliographystyle{ieee}
\bibliography{egbib}
}

\end{document}